\documentclass[sigconf]{acmart}
\renewcommand\footnotetextcopyrightpermission[1]{}
\usepackage{graphicx}
\usepackage{tabularx}
\usepackage{appendix}
\usepackage{amsmath}
\usepackage{algpseudocode}
\usepackage{fancyhdr}
\usepackage[utf8]{inputenc}
\usepackage{verbatim}

\newtheorem{definition}{Definition}
\usepackage[linesnumbered,ruled,vlined]{algorithm2e}
\newtheorem{example}{Example}

\setcopyright{none}

\title{Fast Falsification of Neural Networks using Property Directed Testing}

\author{Moumita Das}
\affiliation{
  \institution{Ericsson Research}
  \city{Bangalore}
  \country{India}
   }
 \email{moumita.a.das@ericsson.com}
 
\author{Rajarshi Ray}
\affiliation{
  \institution{Indian Association for the Cultivation of Science}
  \city{Kolkata}
  \country{India}
}
\email{rajarshi.ray@iacs.res.in}

\author{Swarup Kumar Mohalik}
\affiliation{
  \institution{Ericsson Research}
  \city{Bangalore}
  \country{India}
    }
\email{swarup.kumar.mohalik@ericsson.com}

\author{Ansuman Banerjee}
\affiliation{
  \institution{Indian Statistical Institute}
  \city{Kolkata}
  \country{India}
}
\email{ansuman@isical.ac.in}
\begin{abstract}
Neural networks are now extensively used in perception, prediction and control of autonomous systems. Their deployment in safety critical systems brings forth the need for verification techniques for such networks. As an alternative to exhaustive and costly verification algorithms, lightweight falsification algorithms have been heavily used to search for an input to the system that produces an unsafe output, i.e., a counterexample to the safety of the system. In this work, we propose a falsification algorithm for neural networks that directs the search for a counterexample, guided by a safety property specification. Our algorithm uses a derivative free sampling based optimization method. We evaluate our algorithm on 45 trained neural network benchmarks of the ACAS Xu system against 10 safety properties. We show that our falsification procedure detects all the unsafe instances that other verification tools also report as unsafe. Moreover, in terms of performance, our falsification procedure identifies most of the unsafe instances faster, in comparison to the state-of-the-art verification tools for feed-forward neural networks such as NNENUM and Neurify and in many instances, by orders of magnitude. 
\end{abstract}

\begin{document}

\maketitle
\thispagestyle{empty}



\section{Introduction}

\noindent
From simple day-to-day tools like text prediction in emails and messages, to sophisticated auto-pilot systems in modern planes, almost every aspect of our life today involves systems that learn automatically from data for synthesizing optimal control policies. Indeed, recent advances in Machine Learning (ML) and in particular, in the areas of Reinforcement Learning (RL) and Deep Neural Networks (DNN), has made it possible to achieve exceptional sophistication and performance in a wide variety of domains including chip design, image classification, software product lines, resource allocation, scheduling, and controller synthesis. In recent times, both the software and hardware design industry are seriously considering the possibility of including neural components inside their design artifacts, even inside critical system software and hardware, replacing their age-old hand-coded counterparts.  However, in spite of phenomenal research advances and hardware sophistication, these components still pose a plethora of risks towards widespread deployment. These range from privacy concerns, algorithmic bias and black box decision making, to broader questions of hardware alignment, self-improvement, and risk from unexplainable intelligence. Correctness of these systems is thus of paramount concern and needs to be rigorously verified. Given the scale and complexity of today’s system designs and applications, guaranteeing satisfaction of safety objectives for ML designs under all possible input scenarios is a difficult challenge, due to factors such as non-linearity and non-convexity of the model, high dimensional input spaces, real-valued weights etc. 
As a result, the problem of ML safety verification 
has been at the forefront of verification research in recent times \cite{amodei2016concrete}\cite{blaas2020adversarial}\cite{cardelli2019statistical}\cite{chakraborty2019testing}\cite{cousot1992abstract}\cite{gehr2018ai2}\cite{huang2017safety}\cite{katz2017reluplex}\cite{meel2020testing}\cite{pulina2012challenging}\cite{pulina2010abstraction}\cite{scheibler2015towards}\cite{stanforth2018dual}\cite{wu2020robustness}.\\ 

\noindent In this paper, we address the verification problem of feed-forward neural networks with general activation functions. In particular, given a feed-forward neural network and a property, we propose an efficient falsification algorithm that attempts to search for an input to the network that violates the property and thereby proving the network to be unsafe. Our algorithm uses a derivative free sampling based optimization method to direct the search for a falsifying input based on the safety property.
We refer to a property refuting input as a \emph{counterexample}. This kind of a procedure is an archetype of the \emph{falsification} method of testing a system, which has been heavily used in system verification as an alternative to exhaustive and costly verification algorithms that rather attempt to prove the safety of the system at hand~\cite{kleiner2020falsification}\cite{qin2019automatic}. 
Naive falsification techniques such as random testing \cite{pei2021dynamic}\cite{mao2020adaptive} do not generally learn and infer knowledge from the earlier failed test trials on the system. As a result, a large part of the input-space may have to be explored to find a counterexample. In contrast, falsification procedures for software and hardware that are either property directed \cite{lindblad2007property, Bartley2002ACO} or that explore the input-space systematically have been shown to be considerably effective and efficient in comparison to random testing \cite{10.1145/1064978.1065036}. In this work, we propose a falsification algorithm for neural networks that not only learns from the failed test executions of the neural network but also efficiently directs the search for a counterexample towards the input-space of interest based on the property at hand. The core of our falsification procedure is a derivative-free sampling-based optimization method \cite{yu2016derivative} that we tailor to our needs. Since our procedure is sampling-based, it is applicable to neural networks with any type of activation function. Our proposed falsification algorithm is sound but not complete. When a falsifying input has been found by our algorithm, it terminates by declaring the network as unsafe and the reported counterexample indeed violates the safety property of the network. However, when it terminates before finding any falsifying input, one cannot guarantee the absence of any falsifying input and consequently, the safety of the network with respect to the property. \\

\noindent 
We evaluate our algorithm on 45 trained neural network benchmarks of the ACAS Xu system against 10 safety properties. Empirically, we show that our falsification procedure detects all the unsafe instances that other verification tools also report as unsafe. In terms of performance, our falsification procedure identifies the unsafe instances orders of magnitude faster in comparison to the state-of-the-art verification tools for neural networks. Therefore, we believe that our falsification algorithm can complement the process of neural network verification by rapidly detecting the unsafe instances and henceforth directing the effort of verification on the rest of the instances with sound and complete algorithms. As a result, the overall time to verify a set of instances can be drastically reduced by adopting our method.   \\


\noindent
The rest of the paper is organized as follows. Section~\ref{rel} presents some related work on neural network verification. Section~\ref{pre} presents the problem definition for this work. The detailed method and algorithms of this work are explained in Section~\ref{method}. Section \ref{result} discusses the performance comparison of our method with other existing tools, while Section~\ref{conc} concludes this work.

\section{Related Work} \label{rel}

The integration of neural networks as perception and control components in safety critical systems demands their formal verification. The verification problem of neural networks for even simple properties is known to be NP-complete \cite{katz2017reluplex, katz2019marabou}. Several complete and incomplete algorithms have been proposed for the verification of feed-forward neural networks in the recent literature. The algorithms can be broadly classified into three categories: (i) methods that reduce the verification problem to the feasibility of a mixed integer linear program (MILP) \cite{DBLP:journals/corr/LomuscioM17, bastani2017measuring}, (ii) methods that reduce the verification problem to the satisfiability of an SMT formula \cite{katz2017reluplex, katz2019marabou} and (iii) methods based on geometric set propagation to represent all possible outputs of the network \cite{bak2020improved, tran2020nnv}. Algorithms in the first two categories are complete. In the third category,
algorithms may trade-off completeness for performance by over-approximating the possible outputs of the neural network.  NNENUM~\cite{bak2020improved} is a recent tool that proposes a complete geometric set propagation algorithm for verification of neural networks with ReLU activation function. The tool has successfully verified all the ACAS Xu benchmark instances with a significant performance improvement in comparison to some other tools such as NNV \cite{tran2020nnv} and Marabou \cite{katz2019marabou}. The Neural Network Verification tool (NNV)~\cite{tran2020nnv} can perform both exact and over-approximate reachability analysis with a particular focus on the verification of closed-loop neural network control systems. NNV uses set representations such as polyhedra, zonotopes and star sets that allows for a layer-by-layer computation of the exact reachable set for feed-forward deep neural networks with ReLU activation function. A symbolic interval propagation with an adaptive node splitting strategy has been presented in the tool Verinet ~\cite{henriksen2019efficient} to verify robustness properties of feed-forward neural networks. The verification algorithm is sound and complete for networks with ReLU activation and sound for networks with sigmoid and tanh activation functions. A gradient-descent based counterexample search algorithm is also included in the tool. ETH Robustness Analyzer for Neural Networks (ERAN)~\cite{singheran} is a state-of-the-art sound, precise, scalable, and extensible analyzer that automatically verifies safety properties of neural networks with feedforward, convolutional, and residual layers against input perturbations.  The properties include proving robustness against adversarial  perturbations based on changes in pixel intensity, geometric transformations of images and more. It is based on abstract interpretation and has been used for verification of MNIST, CIFAR-10, and ACAS Xu benchmarks. An efficient method for the verification of ReLU-based feed-forward neural networks that outperforms many of the state-of-the-art tools is proposed in \cite{DBLP:conf/aaai/BotoevaKKLM20} and is implemented in the tool Venus. The algorithm exploits dependency relation between the hidden-layer nodes for pruning the search space of the MILP obtained from the network. Symbolic interval propagation and input domain splitting techniques are augmented in addition. However, the comparison of Venus with other tools shows that Neurify \cite{DBLP:conf/nips/WangPWYJ18} is the fastest in finding counterexamples. 

In this paper, we perform neural network verification in a different way. We apply a sampling-based falsification method that can rapidly detect the unsafe instances (neural networks together with properties) and thus direct the effort of complete verification on the rest of the instances. In particular, our falsification algorithm uses property directed derivative-free sampling to find the falsifying inputs by shrinking the search space. As a result, the overall time to verify a set of instances is drastically reduced. We show that our algorithm outperforms the state-of-the-art tools such as Neurify and NNENUM in counterexample generation on the ACAS Xu benchmarks.


\section{Preliminaries and Problem Definition} \label{pre}
In this section, we present the background relevant to our work and the problem definition. We begin with the definition of a feed-forward neural network and the associated falsification problem. We then discuss a sampling-based derivative free optimization algorithm that forms the core of our proposed falsification algorithm for neural networks. \\

\noindent 
A feed-forward neural network consists of a finite set of nodes, called neurons, arranged in finitely many layers. Every neuron produces a value called its \emph{activation}. The activation of a neuron in a layer is propagated to every neuron in the immediate successor layer by means of weighted connections. The activation $a_j$ of a neuron $j$ in a layer is computed sequentially in two steps. The first step consists of computing the weighted sum of the activations received from the neurons in the preceding layer together with the addition of a constant bias $b_j$ associated with the neuron, i.e., $g_j = b_j + \sum_{i=1}^p w_{i,j}*a_{i}$, where $w_{i,j}$ is the weight of the link connecting the neuron $i$ of the preceding layer and $p$ denotes the number of neurons in that layer. Next, an activation function $act: \mathcal{R} \to \mathcal{R}$ which can be potentially non-linear, is applied on the weighted sum to get the activation $a_j = act(g_j)$. Few of the common activation functions are \emph{ReLU}, \emph{Sigmoid} and \emph{Tanh}. The first layer of the network with no preceding layer is the input layer, the last layer is the output layer and the intermediate layers are called the hidden layers of the neural network. A feed-forward neural network with $n$ and $m$ neurons in the input and output layer respectively represents a non-linear function $f: \mathcal{R}^n \to \mathcal{R}^m$.
\noindent We denote the activation of the $m$ output neurons with variables, namely $o_1, o_2, \ldots, o_m$. \\

\noindent Given a neural network $f: \mathcal{R}^n \to \mathcal{R}^m$, we consider a safety property to consist of a subset of the domain of the network $\mathcal{D} \subseteq \mathcal{R}^n $ along with a first order logic predicate $\phi$ on the output variable(s) of the network. We assume that the $\mathcal{D}$ specified in a safety property is of the form [$x_1^{\ell}$, $x_1^{u}$] $\times$ [$x_2^{\ell}$, $x_2^{u}$] $\times \ldots $[$x_n^{\ell}$, $x_n^{u}$], where $x_i^{\ell}, x_i^{u}$ represent the respective lower and upper bounds on the input $x_i$ of the network and $\times$ signifies a cross-product. A lower bound can be $-\infty$ and the upper bound can be $+\infty$ as well. The predicate $\phi$ over the output variables of the neural network is 
defined as a boolean combination of arithmetic relations as defined by the following grammar:
\begin{equation}
    \begin{split}
    \phi & := \phi \bowtie \phi | (\phi)  | rel \\
    rel & := var \triangleleft var | var \triangleleft c \\
    var & := o_1 | o_2 | \ldots o_m \\
    \bowtie &:= \lor | \land \\
    \triangleleft &:= \leq | \geq | < | > \\
    \end{split}
\end{equation}

\noindent
where $o_1,o_2,\ldots,o_m$ are the output variables of the neural network, $c$ denotes a real constant. An evaluation of a predicate $\phi$ to either \emph{true} or \emph{false} is obtained from an m-tuple output of a neural network by substituting the values from the m-tuple in place of the corresponding variables $o_i$ and under the usual interpretation of the relational and logical operators in $\triangleleft$ and $\bowtie$ respectively. 

\begin{definition}
Given a feed-forward neural network $f:\mathcal{R}^n \to \mathcal{R}^m$, a predicate $\phi$ on the output variables of the network, and an input $x \in \mathcal{R}^n$, an evaluation of $\phi$ on the output $f(x) \in \mathcal{R}^m$ of the network is denoted as $\phi_{eval}(f(x))$.
\end{definition}

\begin{definition} Given a feed-forward neural network $f: \mathcal{R}^n \to \mathcal{R}^m$ and a safety property consisting of $\mathcal{D} \subseteq \mathcal{R}^n$ together with a predicate $\phi$ on the output variables of $f$, the network is said to be safe with respect to the given safety property if and only if $\phi_{eval}(f(x)) = true$, $\forall x \in \mathcal{D}$.
\end{definition}
\noindent We now state the \emph{falsification problem} of a neural network which we address in this paper.

\begin{definition}
Given a feed-forward neural network $f: \mathcal{R}^n \to \mathcal{R}^m$ and a safety property consisting of $\mathcal{D} \subseteq \mathcal{R}^n$ together with a predicate $\phi$ on the output variables of $f$, the falsification problem is to search for an $x \in \mathcal{D}$ such that $\phi_{eval}(f(x))=false$.
\end{definition}

\section{Methodology}\label{method}

\noindent We now present the details of our falsification algorithm for neural networks. 
Given a neural network and a safety property, our proposed falsification algorithm not only learns from the observed test executions of the neural network but also efficiently directs the search for a counterexample towards the input-space of interest based on the property at hand. The directed search is achieved by casting an optimization problem from the given property $\phi$ that we intend to falsify. We first present the details of this construction.

\subsection{Framing Optimization Problem from the Property}
The key to our falsification algorithm is a heuristic that directs the search for a counterexample based on the given safety property of the neural network that is to be falsified. For an illustration of the heuristic, consider the simple case when $\phi$ is only a term with $o_i \triangleleft o_j$. Based on the type of the relation, the decision to either construct a maximization or a minimization problem is made. If the relation is $\leq$ or $<$, our idea is to search for samples in the domain of the neural network that maximizes the variable $o_i$ so that we find a sample for which the relation $o_i \triangleleft o_j$ evaluates to false. Alternatively, we may search for samples that minimizes $o_j$ in order to have a false evaluation of $\phi$. In this way, we intend to direct the search of inputs in the domain that drives the output of the network towards the \emph{boundary} separating the unsafe and safe region and thereafter, looking for inputs for which the network's output \emph{crosses over} from the safe to the unsafe region. This directed searching can be achieved with the help of state-of-the-art solvers by means of solving either one or both of the following optimization problems, as relevant to the property context:

\begin{equation}
maximize \textit{ } o_i \textit{ s.t. } x \in \mathcal{D}, minimize \textit{ } o_j \textit{ s.t. } x \in \mathcal{D}.
\end{equation}
\noindent We similarly address for the $\geq$ or $>$ relation. When $\phi$ consists of many terms joined together with logical connectors, our heuristic iterates over the terms one at a time and frames the optimization problem as discussed, and invokes the solver. For some special structures of $\phi$, the heuristic constructs the optimization problem as shown in Table \ref{tab:heuristicOpt}.

\begin{table*}[ht]
\begin{center}
{
\begin{tabular}{|p{9cm}|p{4cm}|}
\hline
\bf{$\phi$} & \bf{Optimization Problem} \\
\hline
 ($o_i \leq o_{j1}$) $\bowtie$ ($o_i \leq o_{j2}$) $\bowtie \ldots \bowtie$ ($o_i \leq o_{jk}$) & maximize $o_i$ s.t. x $\in \mathcal{D}$ \\
\hline
($o_i \geq o_{j1}$) $\bowtie$ ($o_i \geq o_{j2}$) $\bowtie \ldots \bowtie$ ($o_i \geq o_{jk}$) & minimize $o_i$ s.t. x $\in \mathcal{D}$ \\
\hline
(($o_i \leq o_{j1}$) $\bowtie \ldots \bowtie$ ($o_i \leq o_{jk}$)) $\bowtie$ (($o_p \leq o_{p1}$) $\bowtie \ldots \bowtie$ ($o_p \leq o_{pm}$)) & maximize $o_i$ s.t. x $\in \mathcal{D}$ or maximize $o_p$ s.t. x $\in \mathcal{D}$ \\
\hline
(($o_i \leq o_{j1}$) $\bowtie \ldots \bowtie$ ($o_i \leq o_{jk}$)) $\bowtie$ (($o_p \geq o_{p1}$) $\bowtie \ldots \bowtie$ ($o_p \geq o_{pm}$)) & maximize $o_i$ s.t. x $\in \mathcal{D}$ or minimize $o_p$ s.t. x $\in \mathcal{D}$ \\
\hline
\end{tabular}
}
\end{center}
\caption{\em Choice of Optimization Problem from $\phi$}
\label{tab:heuristicOpt}
\end{table*}

\noindent
The first two entries in the table highlight the case when there is a common output variable across all the terms of $\phi$ related to the other variable / constant of the term with the same relational operator. In such a structure, the choice taken is to minimize or maximize this variable depending on the relational operator. The last two entries highlight the case when the first two structures repeat, connected with logical connectors, when either one or both the optimization problems can be solved to find a falsifying input. Our heuristic can be extended with other special structures which we plan to explore and experiment as a future work.

\subsection{Classification Based Derivative Free Optimization}
\begin{algorithm}
\SetAlgoLined
\textbf{Inputs:} \\
$f$: A neural network with $n$ input nodes and $m$ output nodes\\
Input node intervals:  $[x_i^{\ell}, x_i^{u}]_{i = 1, n}$\\
Property : $\phi$ \\ 
$\theta$ : stopping condition
\vspace{0.5cm}

optType, $o_{target}$ = AnalyzeSpec($\phi$) \\
$c^*$ = null \\
\While{not Timeout}
{  
   \textbf{/* Sampling */} \\
    \textit{S}, isFalsified = MakeSampleAndEvaluate($f$, $o_{target}$, $\phi$, $[x_i^{\ell}, x_i^{u}]_{i = 1, n}$) \\ 
    
    \If { isFalsified = True} 
    {
        /* Falsifying input found */\\
        Falsifying Input = \textit{S}\\
        Terminate \\
    }
    \If {$(x_i^{u} -x_i^{l}) \leq \theta $ for all input variable $x_i$}
    {
         Terminate \\
    }
    $S = S \cup \{c^*\}$ \\
    \textbf{/* The best sample is selected */} \\
    \eIf{optType == maximization} 
    {
         $c^*$ = $argmax_{c \in S} [f(c),o_{target}]$ \\
    } 
    {
         $c^*$ = $argmin_{c \in S} [f(c),o_{target}]$ \\
    }

    \textbf{/* Learning */} \\
    $[x_i^{l}, x_i^{u}]_{i = 1, n}$ = Learning ($S$, optType)\\
}
\caption{FFN}\label{algo1}
\end{algorithm}

\noindent
Solving the optimization problem with gradient-based routines requires computing the partial derivative of the complex non-linear function that a neural-network $f$ represents. Moreover, for neural networks with non-smooth activation functions such as ReLU, the gradient of $f$ may not be defined everywhere in the domain. We therefore resort to a sampling based derivative-free optimization algorithm. In particular, we use a classification based algorithm RACOS (RAndomized COordinate Shrinking) proposed in \cite{yu2016derivative} since this algorithm learns from the earlier test samples and accordingly shrinks the search space. \\

\begin{algorithm}
\SetAlgoLined
\textbf{Inputs:}\\
$f$: A neural network with $n$ input nodes and $m$ output nodes\\
Input Node Intervals :  $[x_i^{l}, x_i^{u}]_{i = 1, n}$ \\
Property : $\phi$ \\
Target output variable : target\\

\textbf{Outputs:} set of samples \textit{S}\\
\vspace{0.5cm}
\textit{S} = $\emptyset$\\
/*$\rho$: number of samples */\\
\For{i = 1 to $\rho$} 
  {  
    \For{j = 1 to n}
        {
          $input_j^{i}$ = Sampling($x_j^{l}$, $x_j^{u}$) \\
         }
      \eIf  {$\phi_{eval}(f(input^i))$ = false}
      {
         return $\{input^i\}$, True /*$f$ is unsafe, falsifying input found */ \\
      }
      {
      \textit{S} = \textit{S} $\cup  \{input^i\}$
       }
   }
   return $S$, False
 \caption{MakeSampleAndEvaluate }\label{algo2}
\end{algorithm}

\begin{algorithm}
\SetAlgoLined
\textbf{Inputs:}\\
\textit{S} : set of samples\\
Input Node Intervals :  $[x_i^{l}, x_i^{u}]_{i = 1, n}$ \\
positive samples size: $k$ \\
Optimization type : optType\\

\textbf{Outputs:} new input ranges\\
\vspace{0.5cm}
/* Select the $k$ best samples from \textit{S} according to the optimization problem */ \\
$pos$= selectPosSample(\textit{S}, optType, $k$) \\
$neg$ = $S$ $\setminus$ $pos$\\
$x_i$ = Randomly chosen from \{1, 2, \ldots, n\} \\
b = Randomly chosen from $pos$ \\
\For{ each T in neg}
  {  
     \eIf  {$b[x_i]  > T[x_i]$ }
      {
         $x_i^{l}$ = random($T[x_i]$, $b[x_i]$)\\
         
      }
      { $x_i^{u}$= random($b[x_i]$, $T[x_i]$)\\
         
      }
   }
   return ( $[x_i^{l}, x_i^{u}]_{i = 1, n}$)
 \caption{Learning }\label{algo3}
\end{algorithm}

\noindent Our top level algorithm (FFN) performs two major steps- a)
analyzes the specification to constructs a non-linear optimization problem and then b) solves the optimization problem.
In a), it finds the optimization type (maximization or minimization) and the target output variable for which this optimization will be framed by calling a method \emph{AnalyzeSpec}. The optimization type is stored in $optType$ and the output variable is stored in $o_{target}$ (see line 6). Output of this method is in turn given to the optimization-problem solver. Details of this method is stated in the previous subsection (4.1). 
Optimization-Problem Solver includes three main steps - sampling, evaluation and learning as shown in Algorithm \ref{algo1}. Along with these,  In the sampling method (\emph{MakeSampleAndEvaluate}, Algorithm \ref{algo2}), $\rho$ (a parameter) randomly chosen input vectors ar selected from the domain $\mathcal{D}$ following a uniform distribution in each iteration, which we call samples. For each sample, the output of the given network is evaluated as stated in Definition 1 in Section \ref{pre}. When the evaluation is false, a falsifying input has been found and the algorithm terminates by declaring the network as unsafe for this property. In this case, \emph{MakeSampleAndEvaluate} returns to Algorithm FFN, a falsifying input together with a flag \emph{isFalsified} that is set to $True$. Otherwise, it returns the sample set $S$ and the flag \emph{isFalsified} which is set to $False$. The same steps are repeated until either a falsifying input is found or the predefined number of samples ($\rho$) have been generated. The best sample observed by the algorithm across iterations is stored in $c^*$ (see line 22, line 24) and it is always kept as a member of the samples set $S$ (line 19). In line 22 and line 24, the second argument $o_{target}$ to the $argmax$ and $argmin$ function denote the output variable with respect to which the maximizing and respectively the minimizing argument to $f$ is to be taken.\\

\noindent If any falsifying example is found from Algorithm \ref{algo2}, FFN terminates, otherwise it checks the size of the input intervals. When the size of every input interval is less than the predefined threshold $\theta$, FFN terminates. Otherwise, it calls the learning method (\emph{Learning}). In this learning phase (shown in Algorithm \ref{algo3}), the $\rho$ samples in an iteration are segregated into $k$ positive and $\rho-k$ negative samples for a parameter $k$ of the algorithm (see line 8). For a maximization problem, the samples evaluating to the $k$ maximum values of the output $o_{target}$ are considered as \emph{positive samples} and the rest are considered as negative samples. Similarly for a minimization problem, the samples evaluating to the $k$ smallest values of the output are considered to be positive. The search-space is pruned in order to remove every negative sample from the search-space. This is achieved by first selecting an input $x_i$ and a sample $b$ from the set of positive samples $pos$ in random (see line 10 and line 11). Now, for every negative sample $T$ in the set of negative samples $neg$, $T[x_i]$ is compared with $b[x_i]$ (line 12-17). Depending on the result of this comparison, either the lower or the upper bound of the $i$th input $x_i$ is adjusted in order to eliminate the negative sample $T$ from the search-space. If $b[x_i]$ is larger than $T[x_i]$, lower bound of $x_i$ is updated with a random value between $T[x_i]$ and $b[x_i]$. Otherwise, the upper bound of $x_i$ is adjusted with a random value between $b[x_i]$ and $T[x_i]$. In this way, the search-space shrinks in every iteration.\\

\noindent 
The sampling and the learning continues (as shown in Algorithm \ref{algo1}) until one of the stopping conditions is met. Our algorithm has three different stopping conditions - a) it terminates after producing a falsifying input, b) the size of all the input ranges is less than the predefined threshold $\theta$, and c) A timeout is encountered. \\

\begin{figure} \centering
\includegraphics[width=7cm]{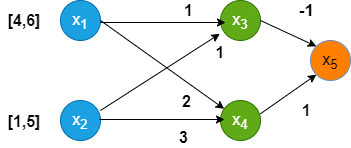}
 \caption{An Example of a Neural Network} 
 \label{fig-ex}
\end{figure}
\begin{example}
Now we discuss the three components of FFN using an example network shown in Figure \ref{fig-ex}. This example network $f$ consists of two inputs $x_1$ and $x_2$ and produces $x_5$ as output.The activation function for each hidden layer node is ReLU. The network has five neurons arranged in three layers.
In this example, we intend to verify the property $\phi$ : $x_5$ < 15 on the domain $\mathcal{D}: x_1 \in [4,6]$ and $x_2 \in [1,5]$. Our heuristic constructs the optimization problem $maximize \textit{ } x_5 \textit{ s.t. } x_1\textit{, }x_2 \in [4, 6] \times [1, 5]$. Algorithm \ref{algo2} samples the domain $\mathcal{D}$ as discussed above. Now consider a run of the algorithm where the first random sample generated is $x_1 = 4$ and $x_2 = 2$. For this sample, $x_5$ now evaluates to 8 and therefore $\phi_{eval}(x_5)$ is true. Evaluation for $x_5$ is done using the formula stated in Definition 1 in Section 3. For this example, consider that $\rho$, the total number of samples observed in each iteration is 3. All the 3 samples in the samples set $S = \{\langle 4,2,8 \rangle, \langle 6,4,14 \rangle, \langle 5,4,13 \rangle\}$ satisfy $\phi$. In this example, we consider the threshold $\theta$ = 0.01 and the number of positive samples $k$ to be 1. Now, Algorithm \ref{algo1} checks the difference in the input ranges for $x_1$ and $x_2$.  Since, the difference for $x_1$ ($|6-4|$) and the difference for $x_2$ ($|5-1|$) are both greater than $\theta$, the algorithm goes for further shrinking of the input ranges. In this illustration, we represent a sample by a triplet where the first two entries denote the value of $x_1$ and $x_2$ respectively and the third entry denotes the evaluation of the neural network $f$ on the corresponding input in the first two entries. As the optimization problem is a maximization type, Algorithm \ref{algo3} segregates positive and negative sample sets as - $pos =\{\langle 6,4,14 \rangle\}$ and $neg = \{\langle 4,2,8 \rangle,\langle 5,4,13 \rangle \}$, since 14 is the maximum valued output of the network. A randomly chosen input dimension $x_i$ = 1 ($x_1$) is taken (see line 9) for comparison of the positive and the negative samples in this dimension and learning from the positive samples. As $pos[x_i] > neg[x_i]$ for the first element of $neg$, the lower bound of the input interval of $x_1^l$ is updated to a random value in the interval [4, 6], say 5. The new input interval for $x_1$ hence becomes [5,6]. For the second element of $neg$, again consider the randomly chosen dimension $x_i$ = 1 ($x_1$). Now, because $pos[x_i] > neg[x_i]$, the new upper bound in the input interval on dimension $i$, i.e., $x_1^l$ is updated to a randomly chosen input from the interval [5, 6] and the new input interval for $x_1$ hence becomes [6,6]. The Learning algorithm returns the shrinked search space [6,6] $\times$ [1,5]. This process continues until one of the stopping condition is encountered.
Now consider that a random sample is generated as $x_1 = 6$ and $x_2 = 5$. For this chosen sample, $x_5 = f(x_1 = 6,x_2 = 5)$ evaluates to 16. Now, $\phi_{eval}(x_5)$ is false and therefore, the sample is a falsifying input. Algorithm \ref{algo1} terminates by declaring the network as unsafe for $\phi$.  
\end{example}

\section{Implementation and Evaluation} \label{result}
\begin{figure}
 \centering
\includegraphics[width=\linewidth]{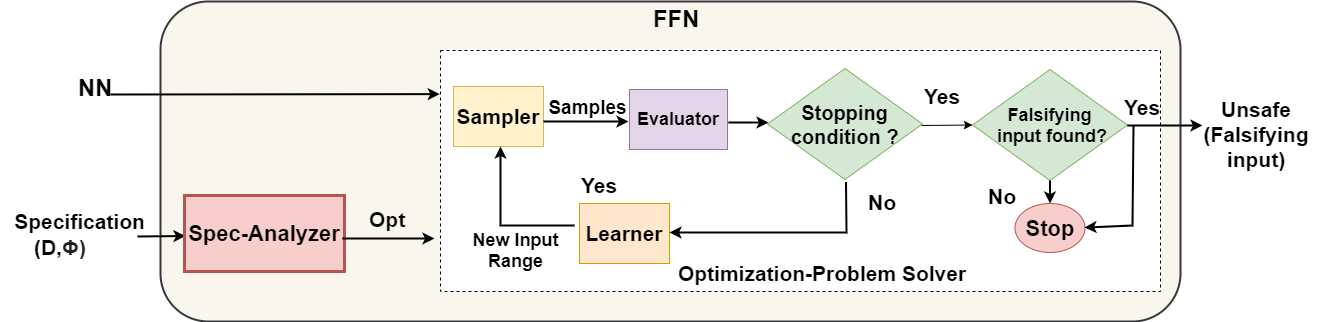}
 \caption{Internal Architecture of FFN Tool} 
 \label{fig3}
\end{figure}

\noindent We implement a tool \emph{Fast falsification of Neural network (FFN)} using the algorithms discussed in Section \ref{method}. The internal architecture of FFN is shown in Figure \ref{fig3}. The inputs to our tool FFN are a neural network ($f$) and a specification ($\mathcal{D} \subseteq \mathcal{R}^n$ and a predicate $\phi$) on the network. If $f$ is unsafe with respect to the given property, FFN produces a falsifying input as an output. The tool consists of two major blocks, a \textbf{Spec-Analyzer} and an \textbf{Optimization-Problem Solver}. \textbf{Optimization-Problem Solver} performs 3 main steps - sampling, evaluation and learning  as discussed in Section \ref{method}. FFN has 3 stopping conditions - a) it terminates after getting a falsifying input, b) it encounters a timeout and terminates and c) the difference between all the input ranges ($x_i^u -x_i^l \le \theta$) is less than the predefined threshold value $\theta$. In this case, we set timeout as 60 secs. Hence, FFN terminates after running for 60 secs if no falsifying input is found. For this experiment we set $\theta$ as $10^{-6}$.\\

\noindent We compare the performance (in terms of execution time) of our proposed tool - FFN with a recent neural network
verification tool - NNENUM \cite{bak2020improved} that uses geometric path enumeration for neural network verification. 
Our experiments are performed on Ubuntu Linux 18.04, 8 GB RAM and an Intel(R) Core(TM) i5-8250U CPU running at 1.60GHz with 8 physical cores. NNENUM reports evaluation on the ACAS Xu benchmark and hence for a comparison, we report the performance of our algorithm on the same benchmark. Implementation of our tool is available at~\cite{fnn}.

\subsection*{ ACAS Xu Benchmarks}
\noindent Airborne Collision Avoidance System X Unmanned (ACAS Xu) is a set of neural network verification benchmarks \cite{katz2017reluplex} which are designed to avoid midair collisions
of aircrafts by issuing horizontal maneuver advisories \cite{marston2015acas}. These fully connected deep neural networks have 8 layers, 5 input nodes ($\rho$, $\theta$, $\psi$, vown, vint), 5 output nodes and 300 ReLU nodes ( 50 neurons in each hidden layer). The 5 outputs nodes are labeled as - Clear of Conflict (COC), Weak Left (WL), Weak Right (WR), Strong Left (SL) and Strong Right (SR). 10 properties were defined on the network encoding safety properties such as if two aircrafts are approaching each other head-on, a turn command will be advised (property 3). The formal definition of all the properties encoded as linear constraints is available in \cite{katz2017reluplex}.\\

\subsection*{Evaluation}
BNFs of 10 ACAS Xu properties, corresponding target output variable and optimization problem are shown in Table \ref{tab1:tab3}. Here, property 7 is defined with two output variables (SR and SL) as "the scores for “strong right” and “strong left” are never the minimal scores", which is written in BNF as "(SR $>$ COC $\land$  SL $>$ COC) $\lor$ (SR $>$ WL $\land$  SL $>$ WL) $\lor$ (SR $>$ WR $\land$  SL $>$ WR)".
We consider only SR as a target output label (first var in the BNF) for which we frame maximization (first $\triangleleft$ in the BNF) as an optimization problem . \\

\begin{table*}[ht]
\begin{center}
{
\tiny
\begin{tabular}{|c|c|c|c|c|c|}

\hline
\bf{Property} & \bf{Property} & \bf{Input}& \bf{Predicate $\phi$} & \bf{Target} & \bf{Objective}\\
\bf{Number} & \bf{Descriptions} & \bf{domain $\mathcal{D}$}&  & \bf{output} & \\
& & & &  \bf{variable} &\\
& & & &  &\\

\hline
\hline

\hline
& & & & &\\
P1 & The output of COC  &  $\rho$ $\geq$ 55947.691, & COC $\le$ 1500 & COC & Maximization\\
 &  is at most 1500 & vown $\geq$ 1145, vint $\leq$ 60 &  & &\\
& & &  & & \\
P2 & The score for COC & $\rho$ $\geq$ 55947.691, & (COC $<$ SR) $\lor$  (COC $<$ WR) & COC & Maximization\\
 &  is not the maximal score & vown $\geq$ 1145, vint $\leq$ 60 & $\lor$ (COC $<$ SL) $\lor$ (COC $<$ WL) &  & \\
& & & & &\\

P3 & The score for COC  & 1500 $\leq$ $\rho$ $\leq$ 1800,& (COC $>$ WL) $\lor$  (COC $>$ WR) $\lor$ & COC & Minimization\\
&is not the minimal score &  $-$ 0.06 $\leq$ $\theta$ $\leq$ 0.06,  & (COC $>$ SL) $\lor$ (COC $>$ SR) &  & \\
& & $\psi$ $\geq$ 3.10,  & & &\\
& & vown $\geq$ 980, vint $\geq$ 960 & & & \\
& & & & & \\

P4 & The score for COC  & 1500 $\leq$ $\rho$ $\leq$ 1800,& (COC $>$ WL) $\lor$  (COC $>$ WR) $\lor$ & COC & Minimization\\
&is not the minimal score &  $-$ 0.06 $\leq$ $\theta$ $\leq$ 0.06,  & (COC $>$ SL) $\lor$ (COC $>$ SR) &  & \\
& & $\psi$ $=$ 0,  & & &\\
& & vown $\geq$ 1000, 700 $\leq$ vint $\leq$ 800 & & &\\
& & & & & \\
P5 & The score for SR  &  250 $\leq$ $\rho$ $\leq$ 400, & (SR $<$ COC) $\land$ (SR $<$ WL) $\land$  & SR & Maximization\\
 & is the minimal score & 0.2 $\leq$ $\theta$ $\leq$ 0.4,& (SR $<$ WR) $\land$ (SR $<$ SL) &  & \\

 &  & $-$ 3.141592 $\leq$  $\psi$ $\leq$ $-$ 3.141592 $+$ 0.005,  & & &\\
& &   100 $\leq$ vown $\leq$ 400, 0 $\leq$ vint $\leq$ 400 & & &\\
&&&&&\\
P6 & The score for COC  & 12000 $\leq$ $\rho$ $\leq$ 62000, & (COC $<$  WL) $\land$ & COC & Maximization\\
& is the minimal score & (0.7 $\leq$ $\theta$ $\leq$ 3.141592) & (COC $<$ WR) $\land$   &  & \\
& & $\cup$ ($-$ 3.141592 $\leq$ $\theta$ $\leq$ $-$0.7), & (COC $<$ SL) $\land$ (COC $<$ SR) & & \\
& & $-$ 3.141592 $\leq$ $\psi$ $\leq$ $-$ 3.141592 $+$ 0.005,  & & & \\
& &  100 $\leq$ vown $\leq$ 1200, 0 $\leq$ vint $\leq$ 1200 & & & \\
& & &  & &\\

P7 & The scores for SR   &  0 $\leq$ $\rho$ $\leq$ 60760,  & ( (SR $>$ COC) $\land$  (SL $>$ COC) ) $\lor$  &SR & Minimization\\
 &    and SL are never  & $-$ 3.141592 $\leq$ $\theta$ $\leq$ 3.141592, & ( (SR $>$ WL) $\land$  (SL $>$ WL) ) $\lor$  & & \\
 &  the minimal scores & $-$ 3.141592 $\leq$ $\psi$ $\leq$ 3.141592, &( (SR $>$ WR) $\land$  (SL $>$ WR) ) & & \\
& &  100 $\leq$ vown $\leq$ 1200, 0 $\leq$ vint $\leq$ 1200 & & & \\
& & & & &\\

P8 & The score for WL & 0 $\leq$ $\rho$ $\leq$ 60760, & ( (COC $<$ WL) $\land$ (COC $<$ WR) $\land$  & COC & Maximization\\
&   is minimal or the & $-$ 3.141592 $\leq$ $\theta$ $\leq$ $-$ 0.75 ·3.141592,  & (COC $<$ SL) $\land$ (COC $<$ SR) ) $\lor$  &  & \\
& score for COC& $-$ 0.1 $\leq$ $\psi$ $\leq$ 0.1, & ( ( WL $<$ WR) $\land$ (WL $<$ COC) $\land$  & & \\
&  is minimal& 600 $\leq$ vown $\leq$ 1200, 600 $\leq$ vint $\leq$ 1200 & (WL $<$ SL) $\land$ (WL $<$ SR) )& &\\
& & & & & \\
P9 & The score for  & 2000 $\leq$ $\rho$ $\leq$ 7000,  & (SL $<$ COC) $\land$ (SL $<$ WL) & SL & Maximization\\
 & SL is minimal & 0.7 $\leq$ $\theta$ $\leq$ 3.141592,
& $\land$ (SL $<$ WR) $\land$ (SL $<$ SR) &  & \\
& & $-$ 3.141592 $\leq$ $\psi$ $\leq$ $-$ 3.141592 $+$ 0.01,  & & & \\
&& 100 $\leq$ vown $\leq$ 150, 0 $\leq$ vint $\leq$ 150 & & &\\
& & & & & \\
P10 & The score for & 36000 $\leq$ $\rho$ $\leq$ 60760, & (COC $<$  WL) $\land$ (COC $<$ WR) $\land$ & COC & Maximization\\
 &  COC is minimal & 0.7 $\leq$ $\theta$ $\leq$ 3.141592,&  (COC $<$ SL) $\land$ (COC $<$ SR) & & \\
 & & 
$-$3.141592 $\leq$ $\psi$ $\leq$ $-$3.141592 $+$ 0.01,   & & & \\
& & 900 $\leq$ vown $\leq$ 1200, 600 $\leq$ vint $\leq$ 1200 & & & \\
& & & & &\\
\hline
\end{tabular}}
\end{center}
\caption{\em Objective type and target output variable for optimization}   \label{tab1:tab3}
\end{table*}
\noindent 
NNENUM has shown the results for properties 1-4 on all 45 networks. However, they have presented the results for properties 5-10 on only one network.
We observe FFN finds the falsifying inputs for all the instances which NNENUM also declares unsafe. For example, for property 1 on all 45 networks, NNENUM declares safe and FFN could not find any falsifying input within a time bound. Similarly, for property 2, out of 45 networks, NNENUM declares unsafe for 43 cases and for each of these 43 cases FFN also finds falsifying inputs. For properties 3 and 4, both FFN and NNENUM declare only three networks as unsafe. NNENUM tests property 5-10 on a single network only, for a common comparative study, we also test property 5-10 for the same network. We find a falsifying input for property 7 on network 1\_9 and for property 8 on the network 2\_9, NNENUM also declares unsafe for these two properties on the same network. These results show that our tool FFN has the capability to find falsifying inputs and all results match with NNENUM. \\

\noindent 
In Table \ref{tab5:tab5}, we present the execution times for all the instances for which we find a falsifying input. We run FFN for 100 times and list the average execution times of only those instances for which it finds the falsifying input in more than 90\% of the cases. In Table \ref{tab5:tab5}, the 3rd and the 4th column present the number of runs and the number of times it gets the falsifying input respectively. The 5th column of this Table shows the average execution time, while the 6th column shows the number of samples used in each example network for the corresponding property testing. To find the average execution time we consider only those cases for which FFN finds the falsifying input. In each run FFN starts with some random samples and executes all the methods. If it finds any falsifying input it terminates the current run and starts the next run. If no falsifying input is found, it changes the seed value (input to initialize random number generator) and continues the same process until it finds a falsifying input or timeout occurs. The number of samples we choose in every iteration is 30 times the number of input nodes of the network \cite{yu2016derivative}. Hence, for ACAS Xu we take 150 random samples. \\

\begin{figure} \centering
\includegraphics[width=\linewidth]{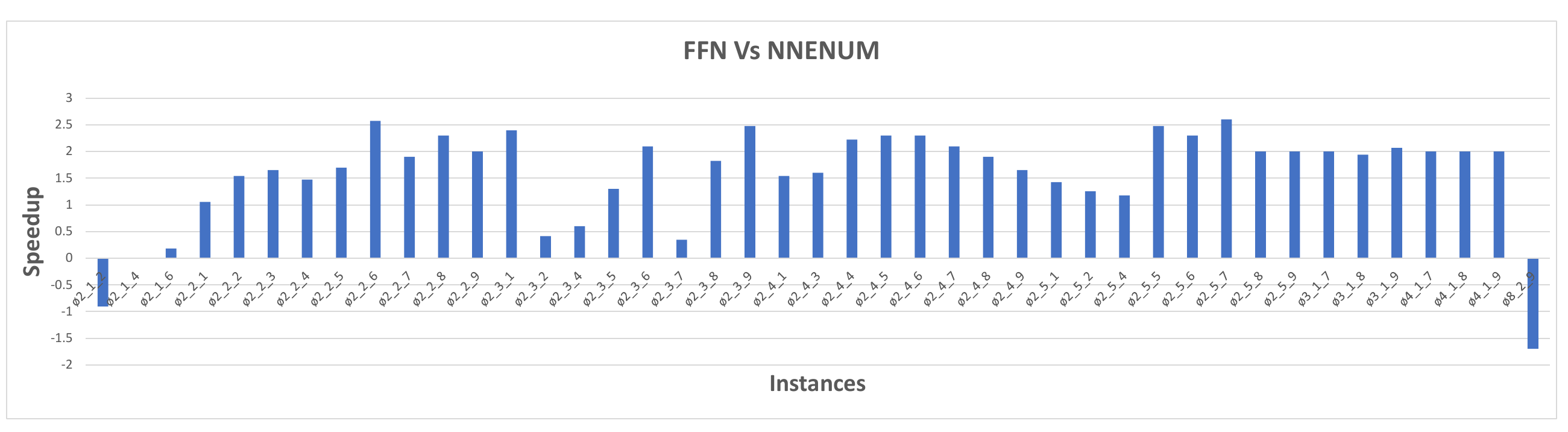}
 \caption{Speedup in falsification: FFN vs. NNENUM} 
 \label{fig4}
\end{figure}

\begin{figure} \centering
\includegraphics[width=\linewidth]{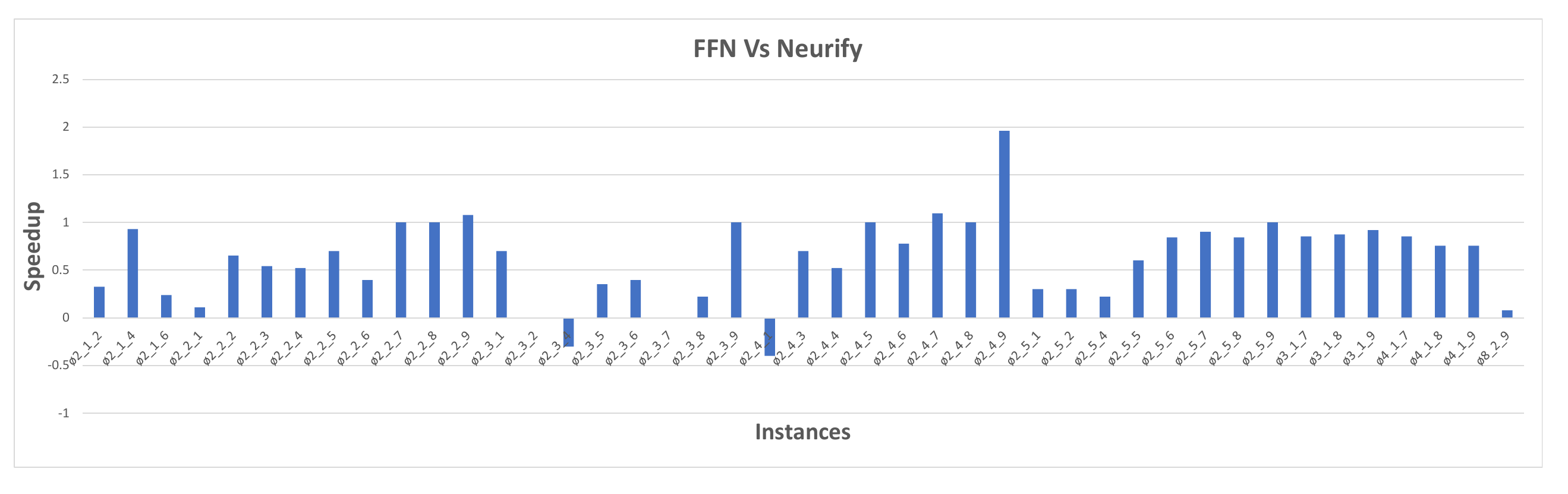}
 \caption{Speedup in falsification: FFN vs. Neurify} 
 \label{fig5}
\end{figure}

\noindent In Table \ref{tab5}, we present a comparative study of FFN, NNENUM and Neurify. In this case, we list only instances for which FFN finds a falsifying input for more than 90\% times (in 100 times total execution). We observe that FFN is significantly faster than NNENUM and Neurify in generating a falsifying input for all most all the cases. In Table \ref{tab5}, 4th and 5th columns show the execution times taken by our tool and NNENUM respectively. In the last column we present the speedup value. 
It is seen that our tool is much faster than NNENUM and Neurify in finding a falsifying input for all most all the instances. For property 2 on network 2\_9 NNENUM and Neurify take 50 secs and 6 secs respectively to find a falsifying input, however FFN can find the falsifying input in .5 second. For only 2 instances ( property 2, networks 3\_4 and 4\_1 ) Neurify finds the falsifying input faster than FFN, while for another 2 instances ( property 2, network 1\_2 and property 8 and network 2\_9 ) NNENUM is faster than FFN to find the falsifying input. The relative speedups are visualized on a log scale -  ($ log(NNENUM/FFN)$ and $log(Neurify/FFN)$)- in the Figures \ref{fig4} and \ref{fig5} respectively.\\


\noindent The limitation of FFN is that we can not guarantee that the network is safe when no falsifying input is found within a certain time bound. If we are unable to find one such input, we terminate on encountering one of the stopping criteria with a timeout of 60 secs. FFN takes 4.5 hours on average to run 450 ACAS Xu instances with a timeout of 60 secs. 

\begin{table*}[ht]
\begin{center}
{
\small
\begin{tabular}{|c|c|c|c|c|c|}

\hline
\bf{Properties} & \bf{Networks} & {\bf{Total number}} & {\bf{Number of times}} & {\bf{FFN time}} & {\bf{Number of}} \\
 &  & {\bf{of execution}} & {\bf{falsifying input found}} & (Sec) & {\bf{samples}} \\
\hline
\hline
P2 & 1\_2 & 100& 100&8 &35872 \\
P2 & 1\_4  &100&100 & 2&10959\\
P2 & 1\_6  &100 &96 &19 &69740\\
P2 & 2\_1 &100 &100 & .07&318\\
P2 & 2\_2  &100 &100 & .02& 100\\
P2 & 2\_3  &100 & 100& .02&99\\
P2 & 2\_4  &100& 100& .03&170\\
P2 & 2\_5  &100 & 100& .02&88\\
P2 & 2\_6  &100& 100& .04& 148\\
P2 & 2\_7  &100 & 100&.01 &36\\
P2 & 2\_8  &100 & 100& .01&75\\
P2 & 2\_9 &100&100 &.5 &2012\\
P2 & 3\_1  &100& 100&.02 &90\\
P2 & 3\_2  &100 & 100&5 &18921\\
P2 & 3\_4 &100&100 & .2&874\\
P2 & 3\_5 &100& 100&.04 &180\\
P2 & 3\_6 &100& 100& .04&148\\
P2 & 3\_7  &100 & 100& .9&3075\\
P2 & 3\_8  &100& 100&.06 &185\\
P2 & 3\_9  &100& 100& .01&35\\
P2 & 4\_1  & 100&100 & .2&986\\
P2 & 4\_3  & 100& 100&.02 &122\\
P2 & 4\_4  & 100& 100& .03&171\\
P2 & 4\_5 & 100& 100& .01&61\\
P2 & 4\_6  &100& 100& .01&52\\
P2 & 4\_7  &100& 100& .008&44\\
P2 & 4\_8  &100& 100& .01&80\\
 P2 & 4\_9  &100&100 & .6&2650\\
P2 & 5\_1 &100& 100& .03&170\\
P2 & 5\_2  &100& 100& .05&253\\
P2 & 5\_4  &100& 100&.06 &273\\
P2 & 5\_5  &100& 100& .02&82\\
P2 & 5\_6  &100& 100& .01&90\\
P2 & 5\_7  &100& 100& .01&46\\
P2 & 5\_8  &100& 100& .01&48\\
P2 & 5\_9&100& 100&.01 &71\\
P3 & 1\_7 &100& 100& .007&1\\
P3 & 1\_8  &100& 100&.008 &1\\
P3 & 1\_9  &100& 100& .006&1\\
P4 & 1\_7  &100& 100& .007&1\\
P4 & 1\_8  &100& 100& .007&1\\
P4 & 1\_9   &100& 100& .007&1\\
P8 & 2\_9 &100& 100& 5&22643\\
\hline
\hline
\end{tabular}}
\end{center}
\caption{\em Tool runtime (sec) to find falsifying input for ACAS Xu instances}   \label{tab5:tab5}
\end{table*}

\begin{table*}[ht]
\begin{center}
{
\small
\begin{tabular}{|c|c|c|c|c|}

\hline
\bf{Properties} & \bf{Networks} & {\bf{FFN}} & {\bf{NNENUM}} & {\bf{Neurify}}  \\
\hline
\hline
P2 & 1\_2 &8 &1&17\\
P2 & 1\_4  & 2&2&17\\
P2 & 1\_6  &19 &29&33\\
P2 & 2\_1 & .07&.8&.09\\
P2 & 2\_2  & .02&.7&.09\\
P2 & 2\_3  & .02&.9&.07\\
P2 & 2\_4 & .03&.9&.1\\
P2 & 2\_5  & .02&1&.1\\
P2 & 2\_6  & .04&15&.1\\
P2 & 2\_7  &.01 &.8&.1\\
P2 & 2\_8  & .01&2&.1\\
P2 & 2\_9 &.5 &50&6\\
P2 & 3\_1 &.02 &5&.1\\
P2 & 3\_2 &5 &13&timeout\\
P2 & 3\_4 & .2&.8&.1\\
P2 & 3\_5 &.04 &.8&.09\\
P2 & 3\_6 & .04&5&.1\\
P2 & 3\_7 & .9&2&timeout\\
P2 & 3\_8 &.06 &4&.1\\
P2 & 3\_9& .01&3&.1\\
P2 & 4\_1 & .2&7&.08\\
P2 & 4\_3 &.02 &.8&.1\\
P2 & 4\_4 & .03&5&.1\\
P2 & 4\_5 & .01&2&.1\\
P2 & 4\_6 & .01&2&.06\\
P2 & 4\_7 & .008&1&.1\\
P2 & 4\_8 & .01&.8&.1\\
 P2 & 4\_9 & .6&27&55\\
P2 & 5\_1 & .03&.8&.06\\
P2 & 5\_2 & .05&.9&.1\\
P2 & 5\_4 &.06 &.9&.1\\
P2 & 5\_5 & .02&6&.08\\
P2 & 5\_6 & .01&2&.07\\
P2 & 5\_7 & .01&4&.08\\
P2 & 5\_8 & .01&1&.07\\
P2 & 5\_9 &.01 &1&.1\\
P3 & 1\_7 & .007&.7&.05\\
P3 & 1\_8 &.008 &.7&.06\\
P3 & 1\_9 & .006&.7&.05\\
P4 & 1\_7 & .007&.7&.05\\
P4 & 1\_8 & .007&.7&.04\\
P4 & 1\_9 & .007&.7&.04\\
P8 & 2\_9 & 5&.1&6\\
\hline
\hline
\end{tabular}}
\end{center}
\caption{\em FFN’s performance to find falsifying input for ACAS Xu instances}   \label{tab5}
\end{table*}

\section{Conclusion}\label{conc}

\noindent In this work, we propose a fast falsification algorithm for feed-forward neural networks. The algorithm performs a property directed search of a counterexample build upon an adaptation of a derivative free, sampling based optimization routine and therefore is applicable on neural networks with general activation functions. The proposed algorithm is sound but incomplete. Evaluation on 45 trained neural network benchmarks of the ACAS-Xu system against 10 safety properties shows that our falsification procedure detects all the unsafe instances that other verification tools also report as unsafe. In terms of performance, our falsification procedure identifies most of the unsafe instances orders faster in comparison to the state-of-the-art verification tools such as NNENUM and Neurify. In many instances, we obtain orders of magnitude speed-up. As a future work, we plan to evaluate our algorithm on adversarial robustness benchmarks such as MNIST and CIFAR-10. 

\noindent 
{\scriptsize
\bibliographystyle{plain}
\bibliography{reference}

\begin{thebibliography}{10}

\bibitem{fnn}
\url{https://github.com/DMoumita/FFN}.

\bibitem{amodei2016concrete}
Dario Amodei, Chris Olah, Jacob Steinhardt, Paul Christiano, John Schulman, and
  Dan Man{\'e}.
\newblock Concrete problems in ai safety.
\newblock {\em arXiv preprint arXiv:1606.06565}, 2016.

\bibitem{bak2020improved}
Stanley Bak, Hoang-Dung Tran, Kerianne Hobbs, and Taylor~T Johnson.
\newblock Improved geometric path enumeration for verifying relu neural
  networks.
\newblock In {\em International Conference on Computer Aided Verification},
  pages 66--96. Springer, 2020.

\bibitem{Bartley2002ACO}
M.~Bartley, D.~Galpin, and T.~Blackmore.
\newblock A comparison of three verification techniques: directed testing,
  pseudo-random testing and property checking.
\newblock In {\em DAC '02}, 2002.

\bibitem{bastani2017measuring}
Osbert Bastani, Yani Ioannou, Leonidas Lampropoulos, Dimitrios Vytiniotis,
  Aditya Nori, and Antonio Criminisi.
\newblock Measuring neural net robustness with constraints, 2017.

\bibitem{blaas2020adversarial}
Arno Blaas, Andrea Patane, Luca Laurenti, Luca Cardelli, Marta Kwiatkowska, and
  Stephen Roberts.
\newblock Adversarial robustness guarantees for classification with gaussian
  processes.
\newblock In {\em International Conference on Artificial Intelligence and
  Statistics}, pages 3372--3382. PMLR, 2020.

\bibitem{DBLP:conf/aaai/BotoevaKKLM20}
Elena Botoeva, Panagiotis Kouvaros, Jan Kronqvist, Alessio Lomuscio, and Ruth
  Misener.
\newblock Efficient verification of relu-based neural networks via dependency
  analysis.
\newblock In {\em The Thirty-Fourth {AAAI} Conference on Artificial
  Intelligence, {AAAI} 2020, The Thirty-Second Innovative Applications of
  Artificial Intelligence Conference, {IAAI} 2020, The Tenth {AAAI} Symposium
  on Educational Advances in Artificial Intelligence, {EAAI} 2020, New York,
  NY, USA, February 7-12, 2020}, pages 3291--3299. {AAAI} Press, 2020.

\bibitem{cardelli2019statistical}
Luca Cardelli, Marta Kwiatkowska, Luca Laurenti, Nicola Paoletti, Andrea
  Patane, and Matthew Wicker.
\newblock Statistical guarantees for the robustness of bayesian neural
  networks.
\newblock {\em arXiv preprint arXiv:1903.01980}, 2019.

\bibitem{chakraborty2019testing}
Sourav Chakraborty and Kuldeep~S Meel.
\newblock On testing of uniform samplers.
\newblock In {\em Proceedings of the AAAI Conference on Artificial
  Intelligence}, volume~33, pages 7777--7784, 2019.

\bibitem{cousot1992abstract}
Patrick Cousot and Radhia Cousot.
\newblock Abstract interpretation frameworks.
\newblock {\em Journal of logic and computation}, 2(4):511--547, 1992.

\bibitem{gehr2018ai2}
Timon Gehr, Matthew Mirman, Dana Drachsler-Cohen, Petar Tsankov, Swarat
  Chaudhuri, and Martin Vechev.
\newblock Ai2: Safety and robustness certification of neural networks with
  abstract interpretation.
\newblock In {\em 2018 IEEE Symposium on Security and Privacy (SP)}, pages
  3--18. IEEE, 2018.

\bibitem{10.1145/1064978.1065036}
Patrice Godefroid, Nils Klarlund, and Koushik Sen.
\newblock Dart: Directed automated random testing.
\newblock {\em SIGPLAN Not.}, 40(6):213–223, June 2005.

\bibitem{henriksen2019efficient}
Patrick Henriksen and A~Lomuscio.
\newblock {\em Efficient Neural Network Verification via Adaptive Refinement
  and Adversarial Search}.
\newblock PhD thesis, Imperial College London, 2019.

\bibitem{huang2017safety}
Xiaowei Huang, Marta Kwiatkowska, Sen Wang, and Min Wu.
\newblock Safety verification of deep neural networks.
\newblock In {\em International conference on computer aided verification},
  pages 3--29. Springer, 2017.

\bibitem{katz2017reluplex}
Guy Katz, Clark Barrett, David~L Dill, Kyle Julian, and Mykel~J Kochenderfer.
\newblock Reluplex: An efficient smt solver for verifying deep neural networks.
\newblock In {\em International Conference on Computer Aided Verification},
  pages 97--117. Springer, 2017.

\bibitem{katz2019marabou}
Guy Katz, Derek~A Huang, Duligur Ibeling, Kyle Julian, Christopher Lazarus,
  Rachel Lim, Parth Shah, Shantanu Thakoor, Haoze Wu, Aleksandar Zelji{\'c},
  et~al.
\newblock The marabou framework for verification and analysis of deep neural
  networks.
\newblock In {\em International Conference on Computer Aided Verification},
  pages 443--452. Springer, 2019.

\bibitem{kleiner2020falsification}
Johannes Kleiner and Erik Hoel.
\newblock Falsification and consciousness.
\newblock {\em arXiv preprint arXiv:2004.03541}, 2020.

\bibitem{lindblad2007property}
Fredrik Lindblad.
\newblock Property directed generation of first-order test data.
\newblock In {\em Trends in Functional Programming}, pages 105--123. Citeseer,
  2007.

\bibitem{DBLP:journals/corr/LomuscioM17}
Alessio Lomuscio and Lalit Maganti.
\newblock An approach to reachability analysis for feed-forward relu neural
  networks.
\newblock {\em CoRR}, abs/1706.07351, 2017.

\bibitem{mao2020adaptive}
Chengying Mao, Xuzheng Zhan, Jinfu Chen, Jifu Chen, and Rubing Huang.
\newblock Adaptive random testing based on flexible partitioning.
\newblock {\em IET Software}, 14(5):493--505, 2020.

\bibitem{marston2015acas}
Mike Marston and Gabe Baca.
\newblock Acas-xu initial self-separation flight tests.
\newblock {\em NASA, Tech. Rep. DFRC-EDAA-TN22968}, 2015.

\bibitem{meel2020testing}
Kuldeep~S Meel, Yash Pote, and Sourav Chakraborty.
\newblock On testing of samplers.
\newblock {\em arXiv preprint arXiv:2010.12918}, 2020.

\bibitem{pei2021dynamic}
Hanyu Pei, Beibei Yin, Min Xie, and Kai-Yuan Cai.
\newblock Dynamic random testing with test case clustering and distance-based
  parameter adjustment.
\newblock {\em Information and Software Technology}, 131:106470, 2021.

\bibitem{pulina2010abstraction}
Luca Pulina and Armando Tacchella.
\newblock An abstraction-refinement approach to verification of artificial
  neural networks.
\newblock In {\em International Conference on Computer Aided Verification},
  pages 243--257. Springer, 2010.

\bibitem{pulina2012challenging}
Luca Pulina and Armando Tacchella.
\newblock Challenging smt solvers to verify neural networks.
\newblock {\em Ai Communications}, 25(2):117--135, 2012.

\bibitem{qin2019automatic}
Xin Qin, Nikos Ar{\'e}chiga, Andrew Best, and Jyotirmoy Deshmukh.
\newblock Automatic testing and falsification with dynamically constrained
  reinforcement learning.
\newblock {\em arXiv preprint arXiv:1910.13645}, 2019.

\bibitem{scheibler2015towards}
Karsten Scheibler, Leonore Winterer, Ralf Wimmer, and Bernd Becker.
\newblock Towards verification of artificial neural networks.
\newblock In {\em MBMV}, pages 30--40, 2015.

\bibitem{singheran}
Gagandeep Singh, Mislav Balunovic, Anian Ruoss, Christoph M{\"u}ller, Jonathan
  Maurer, Adrian Hoffmann, Maximilian Baader, Matthew Mirman, Timon Gehr, Petar
  Tsankov, et~al.
\newblock Eran user manual.

\bibitem{stanforth2018dual}
Robert Stanforth, Sven Gowal, Timothy Mann, Pushmeet Kohli, et~al.
\newblock A dual approach to scalable verification of deep networks.
\newblock {\em arXiv preprint arXiv:1803.06567}, 2018.

\bibitem{tran2020nnv}
Hoang-Dung Tran, Xiaodong Yang, Diego~Manzanas Lopez, Patrick Musau, Luan~Viet
  Nguyen, Weiming Xiang, Stanley Bak, and Taylor~T Johnson.
\newblock Nnv: The neural network verification tool for deep neural networks
  and learning-enabled cyber-physical systems.
\newblock In {\em International Conference on Computer Aided Verification},
  pages 3--17. Springer, 2020.

\bibitem{DBLP:conf/nips/WangPWYJ18}
Shiqi Wang, Kexin Pei, Justin Whitehouse, Junfeng Yang, and Suman Jana.
\newblock Efficient formal safety analysis of neural networks.
\newblock In Samy Bengio, Hanna~M. Wallach, Hugo Larochelle, Kristen Grauman,
  Nicol{\`{o}} Cesa{-}Bianchi, and Roman Garnett, editors, {\em Advances in
  Neural Information Processing Systems 31: Annual Conference on Neural
  Information Processing Systems 2018, NeurIPS 2018, December 3-8, 2018,
  Montr{\'{e}}al, Canada}, pages 6369--6379, 2018.

\bibitem{wu2020robustness}
Min Wu and Marta Kwiatkowska.
\newblock Robustness guarantees for deep neural networks on videos.
\newblock In {\em Proceedings of the IEEE/CVF Conference on Computer Vision and
  Pattern Recognition}, pages 311--320, 2020.

\bibitem{yu2016derivative}
Yang Yu, Hong Qian, and Yi-Qi Hu.
\newblock Derivative-free optimization via classification.
\newblock In {\em Proceedings of the AAAI Conference on Artificial
  Intelligence}, volume~30, 2016.

\end{thebibliography}
}
\end{document}